\documentclass[letterpaper, 10 pt, conference]{ieeeconf}  

\usepackage{xcolor}
\usepackage{gensymb}
\usepackage{cite}

\usepackage{graphicx} 

\title{\LARGE \bf
RainbowSight: A Family of Generalizable, Curved, Camera-Based Tactile Sensors For Shape Reconstruction 
}

\author{
    \authorblockN{Megha H. Tippur and Edward H. Adelson} 
        \authorblockA{Massachusetts Institute of Technology\\
    {\tt\small mhtippur@csail.mit.edu, adelson@csail.mit.edu} 
    } }

\begin{document}

\maketitle
\thispagestyle{empty}
\pagestyle{empty}

\begin{abstract}
Camera-based tactile sensors can provide high resolution positional and local geometry information for robotic manipulation. Curved and rounded fingers are often advantageous, but it can be difficult to derive illumination systems that work well within curved geometries. To address this issue, we introduce RainbowSight, a family of curved, compact, camera-based tactile sensors which use addressable RGB LEDs illuminated in a novel rainbow spectrum pattern. In addition to being able to scale the illumination scheme to different sensor sizes and shapes to fit on a variety of end effector configurations, the sensors can be easily manufactured and require minimal optical tuning to obtain high resolution depth reconstructions of an object deforming the sensor’s soft elastomer surface. Additionally, we show the advantages of our new hardware design and improvements in calibration methods for accurate depth map generation when compared to alternative lighting methods commonly implemented in previous camera-based tactile sensors. With these advancements, we make the integration of tactile sensors more accessible to roboticists by allowing them the flexibility to easily customize, fabricate, and calibrate camera-based tactile sensors to best fit the needs of their robotic systems. 
\end{abstract}

\section{INTRODUCTION}

As roboticists aim to accomplish a wider range of more complex dexterous manipulation tasks, tactile sensors have become a popular choice to improve a robotic end effector’s ability to intelligently interact with its environment. Much like humans, manipulators can greatly benefit from the positional, geometric, and textural information these types of sensors provide. In cluttered or highly dynamic scenes, occlusions to the external, global vision system can occur by other objects or even the manipulator itself. In these situations, it is important for the end effector to be able to obtain local sensory information to complete its task. In some cases, no visual feedback is even available, such as when rummaging in a bag or on the top shelf of a cabinet. Here, the robot must rely solely on its tactile feedback to succeed in its exploration.
\begin{figure}[!ht]
    \centering
    \includegraphics[width=\linewidth]{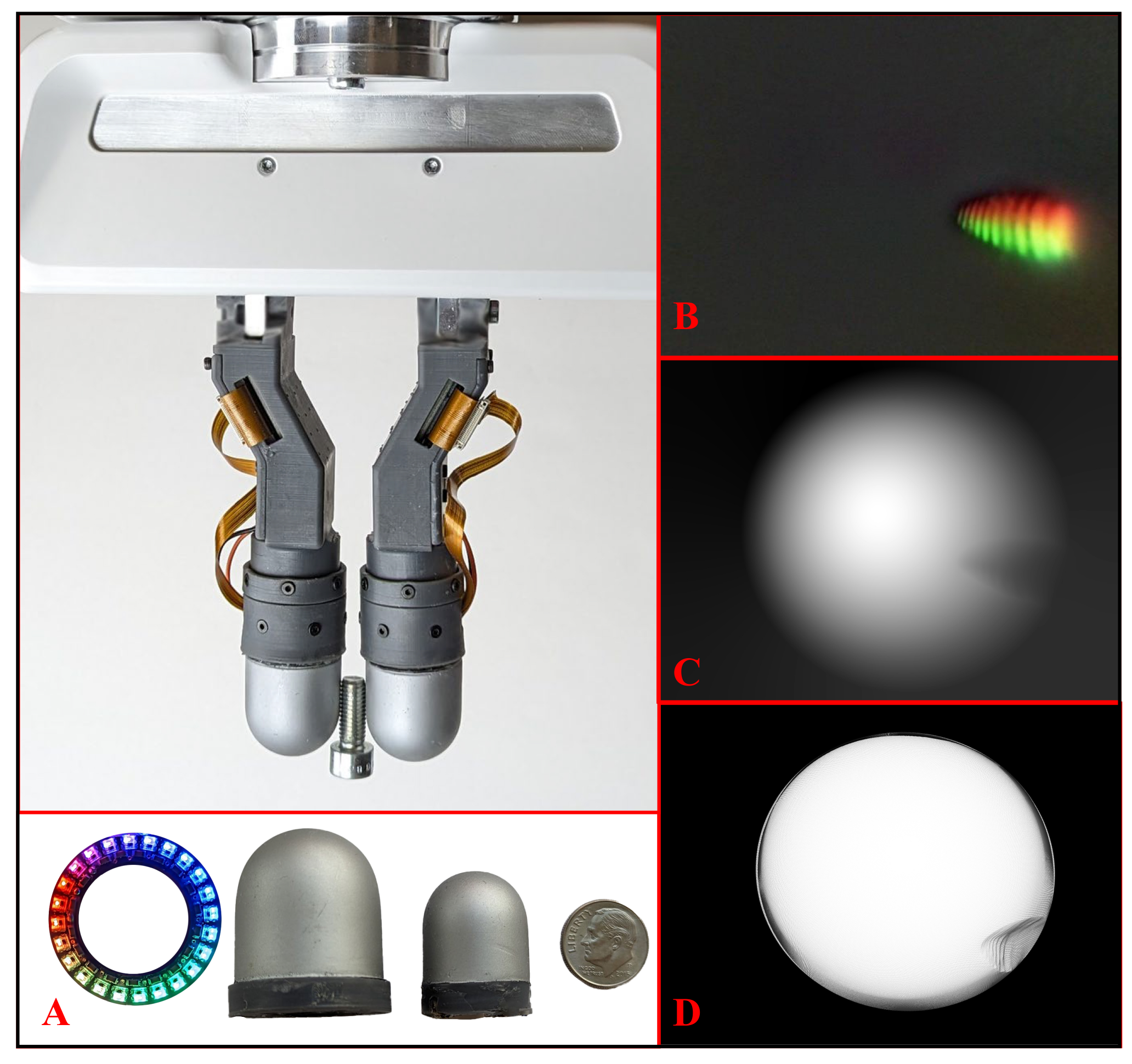}
    \caption{\footnotesize{Two omnidirectional RainbowSight fingers mounted on a parallel-jaw gripper holding an M7 screw. \textbf{(A)} Rainbow illuminated addressable RGB LED ring mounted at the base of the sensors. To the right is an example of a scaled-down version of the sensor with a diameter similar to that of a dime (\(\sim \)20 mm)}. \textbf{(B)} Example difference image captured by camera showing the deformation in the elastomer coating. \textbf{(C)} Depth reconstruction of the sensor surface viewed from sensor base. \textbf{(D)} Point cloud of sensor surface viewed from camera base.}
    \vspace{-20pt}
    \label{fig:teaser}
\end{figure}
A variety of transduction mechanisms can be used to provide tactile feedback, including resistive \cite{sundaram2019learning, fishel2012sensing, ntagios2020robotic}, piezoelectric \cite{zhang2022finger, yong2022soft, tang2019design}, capacitive \cite{zainuddin2015resistive, heyneman2012biologically, muhlbacher2015responsive}, barometric \cite{saloutos2023design, epstein2020bi}, and optical \cite{alspach2019soft, ward2018tactip, lambeta2020digit, shimonomura2019tactile}. In addition, the diversity in size, material composition, and form factor of tactile sensors makes it possible to incorporate them onto different end effector configurations. In particular, camera-based tactile sensors, such as GelSight, have been widely adopted due to their ability to provide high resolution tactile information, such as contact position, object geometries, and applied forces. However, even though much work has been done to obtain accurate contact localization and depth reconstructions with GelSight and GelSight-based sensors, most employ a flat sensing surface with a bulky form factor to obtain these results \cite{yuan2017gelsight, wang2021gelsight, taylor2022gelslim}. Though a flat or curved half-sensing surface may be sufficient for parallel-jaw grippers, end effectors of varying sizes, finger configurations, and DOFs attempting to perform more complicated tasks may not get the most benefit out of camera-based tactile sensors with these shapes. In addition, previous attempts at building curved sensors based on GelSight principles required extensive hardware testing to ensure a proper illumination throughout the entire surface of the sensor, making it difficult to change the shape and size of the sensor. 
We build off our previous work in \cite{tippur2023gelsight360} to introduce RainbowSight, a family of curved, omnidirectional and half-configuration sensors which, to our knowledge, is the first of its kind to utilize a novel rainbow illumination strategy. This strategy allows the user to customize their tactile sensors to fit the size and shape requirements of their specific end effector. Specifically, we show the following contributions and improvements to our previous curved, omnidirectional sensor design: 
\begin{itemize}
  \item A novel rainbow illumination scheme using a semi-specular coating to produce a blended, color gradient desired for photometric stereo techniques. Unlike in \cite{tippur2023gelsight360}, there are no occlusions of the sensor's surface. 
  \item A simplified fabrication procedure utilizing the rainbow illumination scheme that allows customization of shape and size, while maintaining the quality of the final tactile image.
  \item Improved depth reconstructions of the contact deformations occurring on the sensor’s skin. 
\end{itemize}

\section{RELATED WORKS}
Camera-based tactile sensors have proven to be particularly beneficial in many types of manipulation tasks due to their ability to provide high-resolution tactile information by interpreting images of the sensor’s surface deformation when in contact with an object. Many of the current camera-based tactile sensors are based on principles introduced in GelSight, which uses a soft, gel elastomer covered in an opaque, reflective coating to convert geometric and pressure information about a contact region to image data \cite{johnson2009retrographic, yuan2017gelsight}. A camera housed behind the elastomer, along with directionally unique red, green, and blue illumination sources allows photometric stereo methods to be used to estimate the surface normals. These gradient estimates can ultimately  be used with normal integration methods to produce 3D reconstructions of the sensor’s contact deformation \cite{johnson2009retrographic, yuan2017gelsight, wang2021gelsight}. Tactile information, such as depth maps or force sensing, have shown to be useful in many manipulation tasks requiring contact localization \cite{pai2023tactofind, yamaguchi2019recent, lambeta2020digit},  slip detection \cite{li2018slip, yuan2015measurement}, and object geometry estimation \cite{she2021cable, sunil2023visuotactile}.

Much of the initial work on GelSight-based tactile sensors focused on using flat sensing surfaces to improve the spatial resolution, depth reconstruction quality, and force sensing for use in manipulation tasks \cite{yuan2017gelsight, wang2021gelsight, taylor2022gelslim}. More recently, many camera-based tactile sensors with curved, unconventional geometries have been developed to augment a variety of robotic gripper configurations \cite{liu2023gelsight, patel2021digger, tippur2023gelsight360}. Some curved omnidirectional sensors, such as \cite{tippur2023gelsight360} convert traditional GelSight processing methods to work on curved surfaces in order to produce depth maps of the deformed contact region. Other, half-configuration sensors, such as \cite{do2023densetact} use a hemispherical gel dome geometry and collect a variety of different probing shapes to directly translate an RGB difference image to a depth estimation. By adding a randomized pattern to the surface of the gel, the authors were also able to extract 6-axis F/T data from their sensor \cite{do2023densetact}. \cite{althoefer2023miniaturised} uses a force sensitive spring system along with pad-printed markers on the gel’s surface to acquire multi-axis force data and contact geometry. 

Though half sensor configurations provide sufficient sensing area for tasks using parallel-jaw grippers, more dexterous grippers may further benefit from all-around tactile sensors, similar to the curved sensing geometry of human fingertips \cite{lin2021exploratory, pai2023tactofind, andrussow2023minsight, khandate2023sampling}. Due to their large, all-around sensing surface, omnidirectional sensors can sense multiple contact regions, speed up exploration time, prevent damaging collisions, and perform complex in-hand manipulation tasks that would be difficult with flat or even one-sided curved sensors. Many camera-based, omnidirectional tactile sensors have been introduced in recent years. The OmniTact sensor introduced a multi-camera scheme in an elastomer-covered, finger-like shape that was able to estimate the contact angle of objects \cite{padmanabha2020omnitact}. The GelTip sensor shared a similar shape to \cite{padmanabha2020omnitact} and used structured red, green, and blue light with a camera mounted at the base of the internal plastic shell to observe deformations in the sensor skin. Insight, a soft, conically-shaped thumb size sensor, used information from internal texture and structured lighting to train a neural network to estimate contact localization and force estimates within sub-millimeter accuracy \cite{sun2022soft}. Later, a smaller, more fingertip-shaped version of the sensor, Minsight, was introduced and showed the sensor’s performance in a tactile servoing and lump classification task \cite{andrussow2023minsight}. Similarly, the AllSight sensor is able to calculate extremely accurate position and F/T data using a red, blue, green lighting scheme and black markers painted on the elastomer’s surface \cite{azulay2023allsight}. Additionally, they demonstrate zero-shot transferability of their state estimation models to user-manufactured sensors \cite{azulay2023allsight}. Finally, GelSight360 used an LED crossing structure to recreate the unique directionality constraints of the colored lights in order to use photometric stereo techniques to provide high resolution depth reconstructions of the contact regions \cite{tippur2023gelsight360}. In addition, the generalizability of the cross LED illumination scheme was demonstrated by implementing it in a variety of omnidirectional finger shapes \cite{tippur2023gelsight360}. 

In this work, we build on many of the methods introduced in \cite{tippur2023gelsight360} with an alternate illumination scheme to propose a new family of curved tactile sensors of different sizes and shapes. In addition, we make improvements to the sensor hardware and image processing methods to produce more accurate depth reconstructions of deformed regions.

\section{SENSOR DESIGN AND FABRICATION} \label{sensor_design_and_fabrication}

\subsection{Design Criteria}
We focus on introducing a novel illumination system that can be generalized to a variety of curved sensor shapes, along with fitting the sensor and electronics into a compact form factor that can be modularly mounted to a variety of gripper configurations \cite{tippur2023gelsight360, pai2023tactofind}. We improve on GelSight360 by replacing the cross LED structure previously mounted in the body of the sensor with a single rainbow-illuminated addressable RGB LED ring housed at the bottom of the sensor. This prevents occlusions in the sensing surface without sacrificing the integrity of the tactile image. In turn, since the improved illumination scheme is heavily implemented in software rather than hardware, the process to fabricate the sensors is eased, and the sensor sizes and shapes can be scaled down and generalized to account for a variety of curved shapes (including both omnidirectional and half-sensor configurations). This section further describes the illumination system implemented and the fabrication methods used to build the sensors. 
\begin{figure}[!t]
    \centering
    \includegraphics[width=\linewidth]{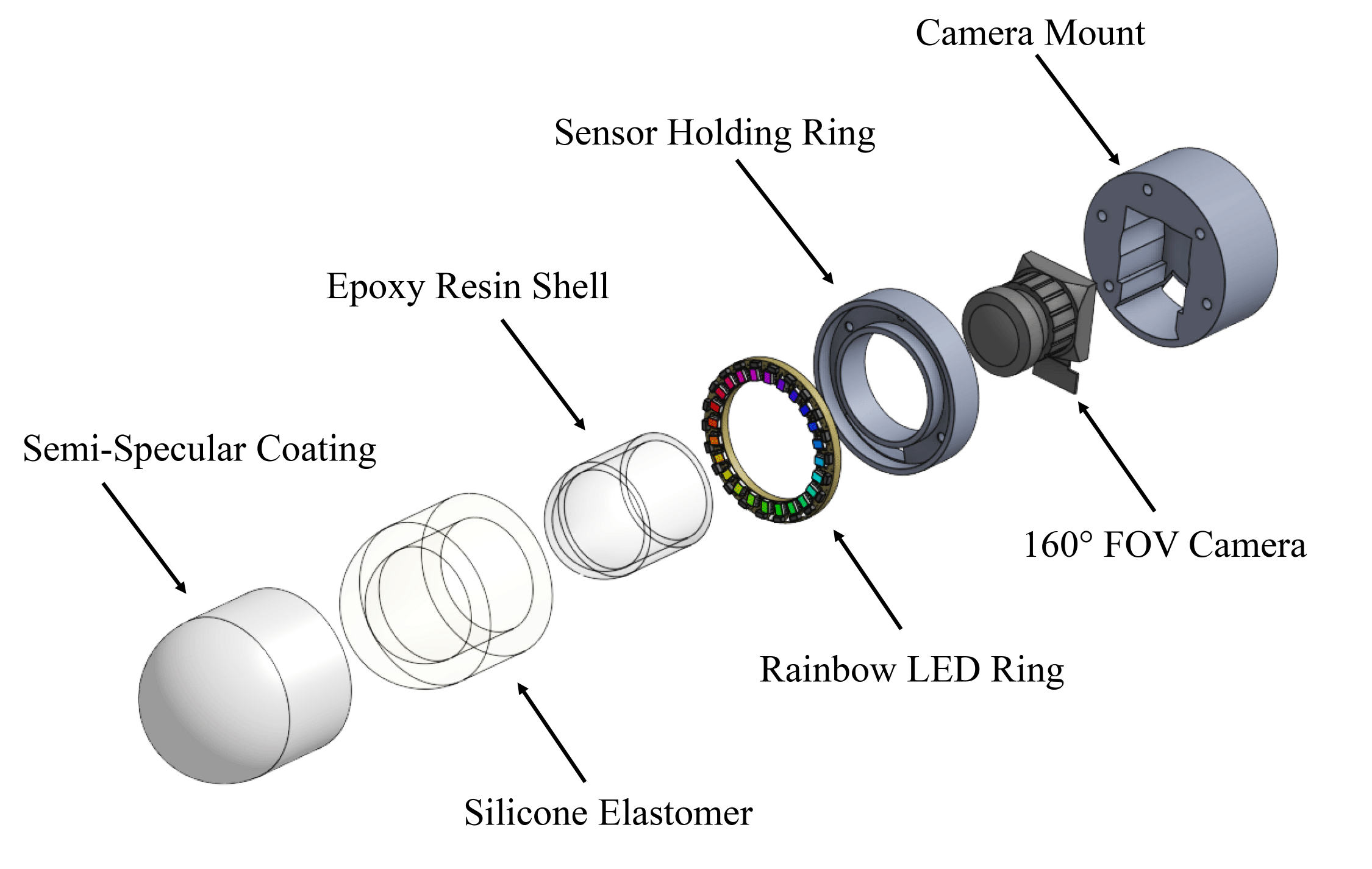}
    \caption{\footnotesize{Exploded view of sensor.}}
    \vspace{-10pt}
    \label{fig:exploded_view}
\end{figure}

\subsection{Sensor Illumination Strategy} \label{sensor_illumination_strategy}
Past flat GelSight and GelSlim sensors use red (\(\sim \)630nm), green (\(\sim \)525nm), and blue (\(\sim \)470nm) light to illuminate a soft, clear gel elastomer coated with an opaque, matte, aluminum-based pigment to capture the gel’s deformation when in contact with an object \cite{johnson2009retrographic}. Because the three colors illuminate the sensor from different directions, photometric stereo techniques can be used to linearly map the RGB intensities collected by a camera housed behind the gel to the surface gradients of the sensor using a look-up-table \cite{johnson2009retrographic, yuan2017gelsight}. These horizontal (Gx) and vertical (Gy) gradients can then be integrated using the Fast Poisson Solver to produce a 3D height map of the deformation of the gel \cite{yuan2017gelsight, wang2021gelsight, taylor2022gelslim}. 

When a semi-specular aluminum coating is used to paint the elastomer (as opposed to the matte, gray Lambertian paint used in \cite{yuan2017gelsight, wang2021gelsight, taylor2022gelslim}), a one-to-one mapping between the RGB intensities and surface reflectance function no longer exists, making it more difficult to estimate the surface gradients. Additionally, due to the reflective properties of the material, the transitions between red, green, and blue are extremely harsh, and do not produce the diffused, blended color gradient observed with Lambertian coated surfaces. However, because of the semi-specular material's sensitivity to smaller changes in the normals, it can actually provide higher resolution signals than the Lambertian case \cite{yuan2017gelsight}. 

We attempt to leverage the high-resolution characteristics of the semi-specular aluminum coating for our curved sensors without sacrificing the desired, smooth, rainbow color gradient observed in sensors using a gray Lambertian coating by utilizing addressable RGB LEDs in our sensors. By weighting the intensity of red, green, or blue light in each LED, the hue around the perimeter of the board appears as a rainbow spectrum. When an object is pressed into the surface of a curved sensor, this blended rainbow gradient, as in Figure \ref{fig:depth_reconstructions}, can be observed when comparing the contact and reference image (later referred to as the difference image). However, for omnidirectional sensors, only the top section of the sensor body is able to achieve this gradient due to the nature of the sensor geometry. Because the light source sits at the base, on the sides of the sensor, the different hues are unable to blend, and the rainbow LEDs act more as colored point sources with the semi-specular coating. Recently, Wang et al. showed that it was possible to leverage deep learning methods to predict relatively accurate gradient estimates and depth reconstructions even when only two colored lights were shined parallel to each other; we employ similar methods to resolve our gradient estimation issues \cite{wang2021gelsight}. 


\subsection{Sensor Fabrication}
By using only a single rainbow LED ring to satisfy the photometric stereo illumination requirements for the sensor, we simplify its fabrication process. Many of the materials and methods used are similar to those in \cite{tippur2023gelsight360} but have been altered to account for improvements in sensor illumination, deformation sensitivity, and overall sensor rigidity. The sensor body consists of three components: 1) a thin, internal resin shell, 2) a transparent, gel elastomer, and 3) an opaque semi-specular coating on the elastomer’s surface. 

\textbf{\textit{Transparent Resin Shell:}} Because of the length and vertically-oriented nature of curved, omnidirectional sensors, an internal, rigid skeleton is often chosen to provide a rigid mounting base and to prevent the gel from delaminating or tearing off\cite{padmanabha2020omnitact, gomes2020geltip, andrussow2023minsight, azulay2023allsight}. For these reasons, we use a 1mm-thick epoxy resin shell as the internal support for the sensors. Smooth-On’s EpoxAcast690 resin is chosen since it produces a transparent, bubble-free shell that is resistant to yellowing caused by UV light over time (unlike other clear SLA resins). In addition, to further the mechanical resiliency of the sensor hardware, we cast the epoxy shell directly into the sensor mounting piece (printed on FormLabs Tough 2000 Resin), rather than gluing it into the base as in \cite{tippur2023gelsight360}. A two-part, gravity molding structure is used to create the epoxy shell. The mold negatives are printed with a FormLabs Form2 SLA printer. After polishing the negatives, Smooth-On MoldStar 20T is used to cast the silicone molds. 

\textbf{\textit{Painted Gel Elastomer:}} Once the epoxy is cured, the shell is removed from the mold using the top covering piece; this piece is re-used when casting the soft gel to ensure the resin shell and elastomer remain aligned on all axes. Unlike in \cite{tippur2023gelsight360}, we chose to couple the base LED ring with the silicone gel rather than the epoxy to minimize ray refractions and reflections that might be caused by polishing imperfections or refractive index mismatches between the resin and clear elastomer. To ensure there is no air-interface between the LEDs and the elastomer, we press-fit the soldered LED ring into the PCB holder in the sensor mount to allow the silicone to cure around the LEDs. The resin shell is primed to promote adhesion to the elastomer. To create the semi-specular coating, a mixture (by weight) of 1 part Print-On Clear Silicone Ink Base (Raw Materials): 0.1 part Print-On Silicone Ink Catalyst: 0.2 part aluminum flake powder: 3 parts of NOVOCS Gloss (Smooth-On) is prepared and sprayed into the mold. A similar procedure described in \cite{tippur2023gelsight360} is used to prepare the soft, clear elastomer.  

\textbf{\textit{Sensor Electronics:}}
We design custom RGB LED PCBs in different ring sizes and shapes to fit the desired footprints of both omnidirectional and half-sensor shapes. Even though RGB LED strips and rings are available off-the-shelf, they come in limited shapes and sizes and are too bulky for our applications. Additionally, we need as many LEDs as spatially possible on the ring in order to successfully mimic a continuous color gradient with the discretized LEDs. Depending on the size of the board being designed, we use either the NeoPixel Addressable 2020 (for larger sensors) or 1515 (for smaller sensors) RGB LEDs with an integrated driver chip (WS2812B or SK6805 respectively). A 0.1\textmu F decoupling capacitor is used for every two RGB LEDs to prevent possible current fluctuations in the IC chip if there are voltage drops or surges. Anywhere between 21 – 28 LEDs are able to be fit on a single side of the various shaped boards.

To control the colors of the addressable LEDs, the ring is connected to an Adafruit\texttrademark{} Pro Trinket  Microcontroller (5V Logic). The microcontroller is chosen due to its small, compact form factor, allowing it to fit in a variety of housing configurations, rather than being housed off the robot entirely since it is best practice to keep the length of the data wire below 1m to minimize possible signal reflections. In addition, when powered with only a microUSB cable, the Pro Trinket has enough onboard memory and current output capability to power up to 5 LED rings at the settings desired for RainbowSight sensors. A 300 ohm resistor is placed on the digital signal line to prevent voltage spikes which could cause communication issues. The FastLED library is chosen to tune the ratios of the red, green, and blue diodes in each of the LEDs since it provides full rainbow HSV color support and brightness controls on the Arduino platform.   


\textbf{\textit{Camera:}} Depending on the sensor size, we chose between two wide-angle cameras to view the entire sensing surface. For larger sensors, we use the 175\degree{} FOV 8MP Sony IMX219 camera, since the inner diameter of the resin shell is large enough to fit its lens. Smaller sensors utilize the 160\degree{} FOV 5MP OV5647 camera since it has a smaller form factor. Both cameras are run off a Raspberry Pi using a CSI connector, and the 640 x 480 images are streamed at 30 fps to a local desktop using mjpg-streamer.  
\section{ALTERNATE SENSOR SHAPES}
\begin{figure}[!t]
    \centering
     \includegraphics[width=\linewidth]
    {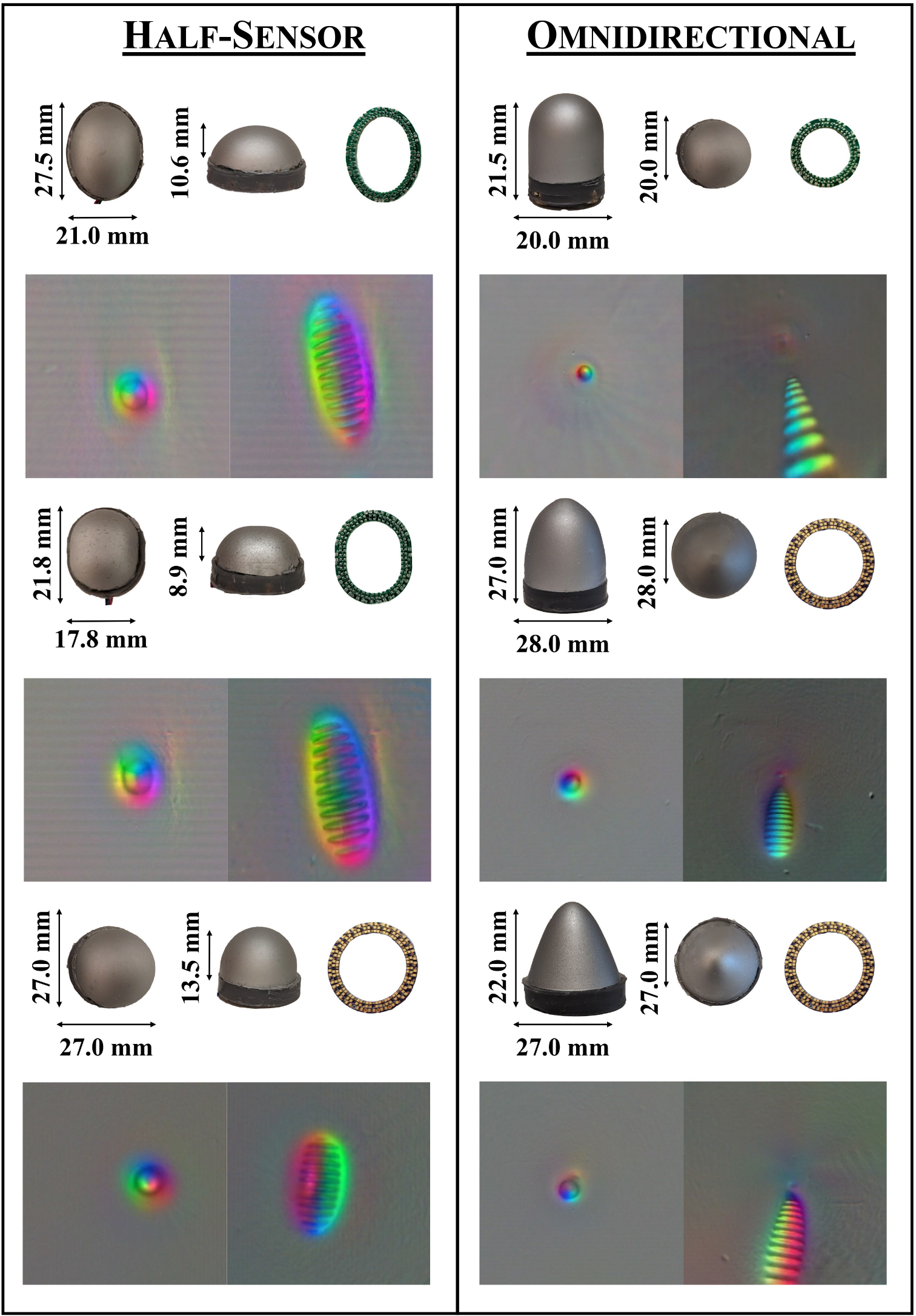}
    \caption{\footnotesize{Example difference images collected when a 4mm ball and M7 screw are pressed into the various sensor shapes and sizes. In the half-sensor configurations and tops of omnidirectional sensors, we are able to achieve the desired rainbow gradient illumination pattern over a large area of the main sensing surface. For omnidirectional sensors, the sensor sides provide adjacent two-color parallel light.
    }}
    \vspace{-21pt}
    \label{fig:alternate_geometry_example}
\end{figure}

In addition to fabricating the sensor shape shown in Figure \ref{fig:teaser}, we demonstrate the novel illumination scheme's generalizability to different sensor geometries (both half-sided and omnidirectional) with the same manufacturing technique described in Section \ref{sensor_design_and_fabrication}. Figure \ref{fig:alternate_geometry_example} shows the sizes of the sensor, the board shape used, and the resulting images produced when a 4mm sphere and M7 screw (respectively) are pushed into each of the sensors. The shapes and approximate sizes of the half-sensor configurations are  inspired by the differing geometries of the commercially available SynTouch BioTac sensors. In these shapes, even though the sensor is not radially symmetric with respect to the vertical axis, the desired blended, rainbow effect is still achieved even though a semi-specular coating is used.  With the removal of the cross LED structure used in \cite{tippur2023gelsight360}, fabricating half-sensor configurations and truly fingertip-sized sensors becomes possible. One drawback of the cross-LED was that because the thickness of the PCBs and LEDs remained constant, as the diameter of the sensor body was reduced, a greater percentage of the sensing area would become severely occluded, limiting the advantages of an omnidirectional sensor. The camera’s perspective model also prevented the cross-LED structure to be properly implemented in half-sensor configurations, since it heavily occluded the main sensing surface due to how close it was to the camera lens. However, since the illumination is now housed at the base of the sensor, a wider range of shapes can be designed without sacrificing either sensing area visibility or the colored illumination’s directionality and blending. 

\section{MEASURING 3D GEOMETRY} \label{measuring_3d_geometry}

\begin{figure*}[!t]
    \centering
    \includegraphics[width=\linewidth]{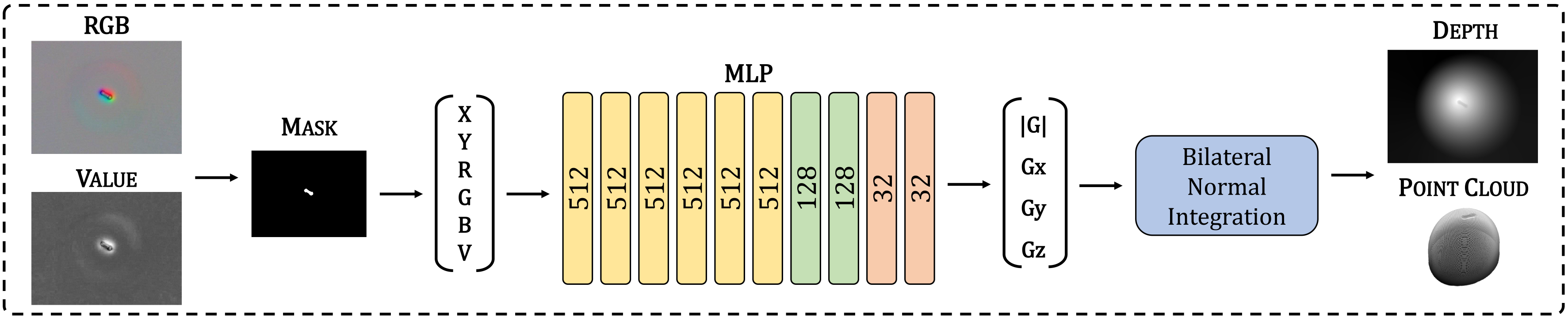}
    \label{fig:depth_pipeline}
\end{figure*}

\begin{figure*}[!t]
    \centering
    \vspace{-20pt}
    \includegraphics[width=\linewidth]{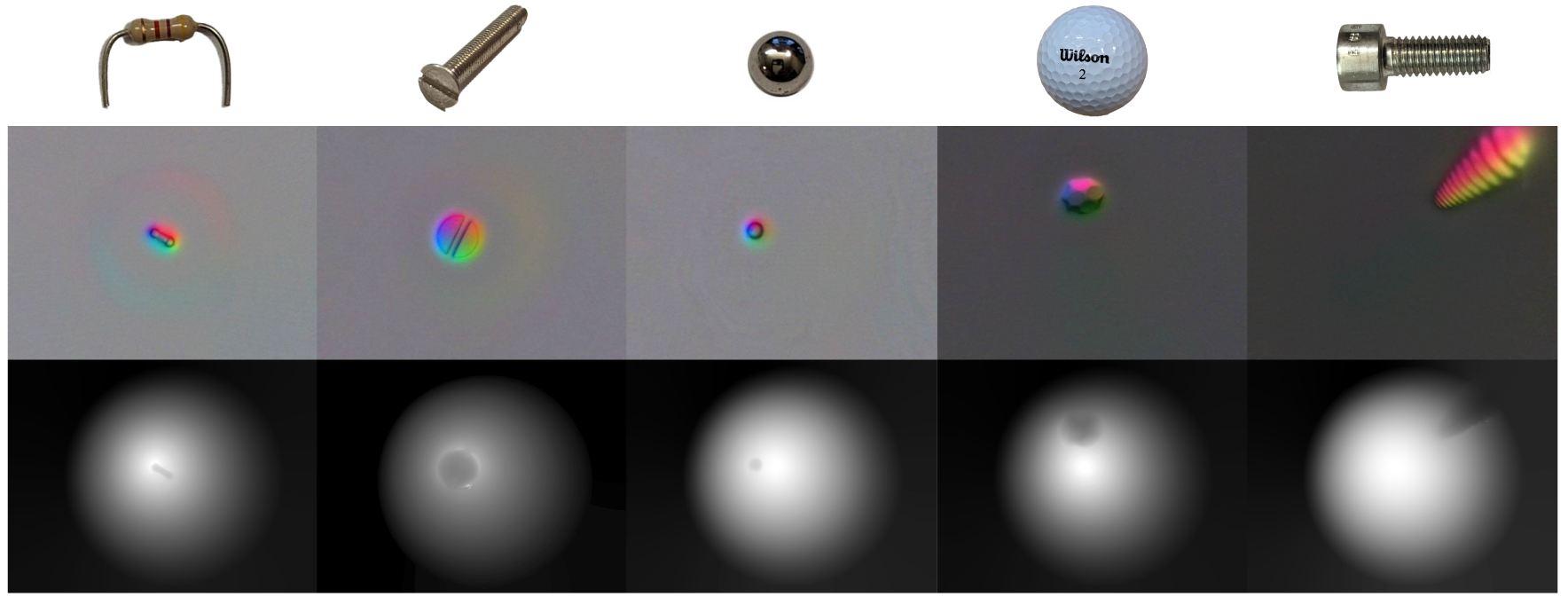}
    \caption{\footnotesize{\textbf{Top Dashed Box:} Depth reconstruction pipeline. \textbf{ Bottom: }Tactile signals collected when different objects are pressed at various locations on the omnidirectional cylinder with hemisphere top sensor shape. \textbf{\textit{Top Row:}} Objects pressed into the sensor surface. \textbf{\textit{Middle Row:}} Tactile difference images of the contact regions. \textbf{\textit{Bottom Row:}} Estimated depth map of the imprinted object in the sensor skin.}}
    \vspace{-17pt}
    \label{fig:depth_reconstructions}
\end{figure*}

Similar to calibration methods used for previous GelSight sensors in \cite{yuan2017gelsight, wang2021gelsight, tippur2023gelsight360}, a lookup table mapping the image’s RGB pixel intensities to the surface gradients of the sensor surface must be constructed. These gradients can then be integrated to reconstruct the deformations in the sensor’s surface. Unlike in \cite{do2023densetact}, which uses an array of different shapes that must be probed into the sensor's surface, our calibration method requires only a spherical probe, making it easier for our calibration method to be generalized to the different shapes and sizes of the sensors tested in this work.

We use a similar calibration and data-processing technique as in \cite{tippur2023gelsight360}, as most of the improved depth reconstruction results come from the illumination improvements, network architecture, and integration methods employed.  The calibration data for the sensor is collected using a tabletop 3020 CNC since it can precisely probe the sensors at the desired positions. The surface mesh of the sensor is sampled at N = 5000 probing points. The CNC autonomously pushes the sphere 4mm into the soft, sensing surface and records images of the deformation.

Because of the variety of sensors possible with this new illumination technique, we aim to implement a universal calibration procedure to collect ground truth data when training the learning-based lookup table, rather than using specified ray-casting methods or coordinate conversions as in \cite{do2023densetact, gomes2020geltip}, respectively. Before the sensor is fully assembled, the fisheye camera is focused to the correct focal length, and a checkerboard calibration is done using OpenCV’s Fisheye Calibration procedure to obtain the camera intrinsic and distortion parameters. With the world coordinates obtained from the CNC probing and the probed sensor images, PnP RANSAC is used to find the camera’s location. Using the camera’s intrinsic and extrinsic parameters, along with the model of the sensor, the gradients of the sensor’s surface and spherical contact from the CNC probing points can be rendered using graphics software. 

The data is prepared for training by first undistorting the images. A difference image is found by subtracting the reference image from the contact image. Thresholding and blob detection are used to mask the probed area, and the pixel coordinates (x, y) along with RGB intensities are collected. The difference image is also converted into the Hue Saturation Value (HSV) color space, and the masked Value layer is collected to create an M x 6 data input for training, where M represents all of the masked pixels from the probed CNC points. The same mask is applied to the rendered normal map to collect the ground truth training data. The gradients are broken into their magnitudes and normalized X, Y, Z components, which is used as the M x 4 ground truth data for training. A multilayer perceptron network (MLP) is trained using ReLU activation with early stopping on M = 1,500,00 points.

The 3D surfaces are reconstructed using normal integration techniques. In previous, flat GelSight sensors, the Fast Poisson Solver was used to integrate the gradients to produce the depth maps \cite{yuan2017gelsight, wang2021gelsight}. However, in \cite{tippur2023gelsight360}, we found that due to the curved nature of our sensor, the Fast Poisson Solver caused distortion in our depth maps due to the lack of discontinuity preservation in the perspective projection case. To address this issue, we utilize Cao et al.’s Bilateral Normal Integration method to generate the depth maps and resulting point clouds \cite{cao2022bilateral}. When integrating a small portion of the gradients, the method can run at about 14Hz. 
\begin{figure}[!t]
    \centering
    \includegraphics[width=\linewidth]{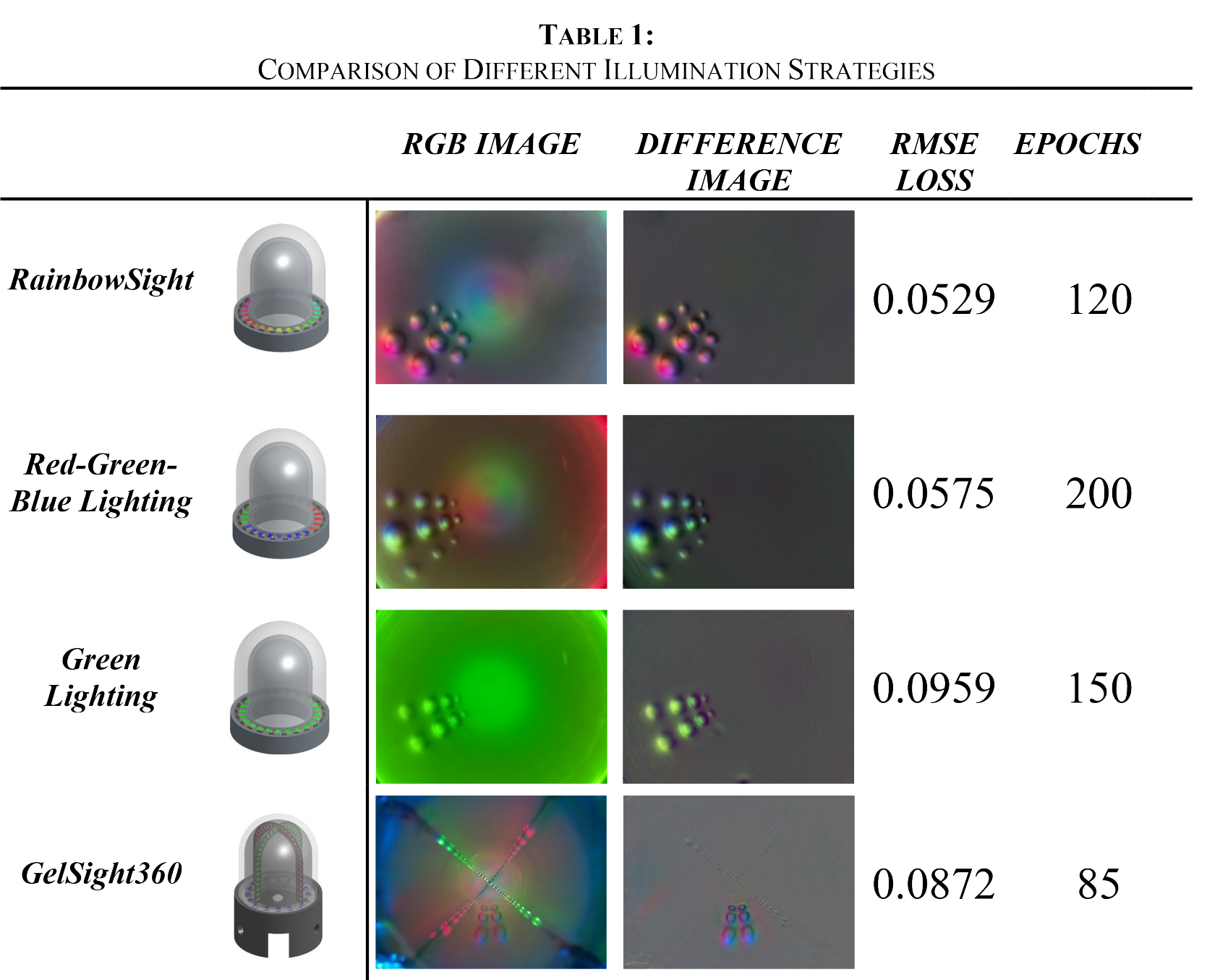}
    \label{fig:comparison_table}
    \vspace{-25pt}
\end{figure}

\textbf{\textit{Results:}} The depth maps generated when objects of different sizes and textures are pressed into the elastomer skin of the calibrated sensor at different locations are shown in Figure \ref{fig:depth_reconstructions}. The general shape and orientation of the object can be recognized with significantly less warping than the results shown in \cite{tippur2023gelsight360}; this can mostly be attributed to 1) a more accurate gradient estimation network and 2) the use of a different normal integration technique. Although the finer details, such as the threads on the screw or the individual textures of the golfball, are lost due to the inherent smoothing applied by this integration technique, the pose and contact location are preserved. 

\begin{figure}[!t]
    \centering
    \vspace{-5pt}
    \includegraphics[width=\linewidth]
    {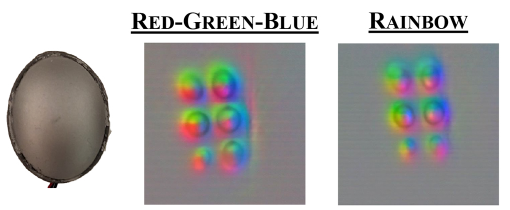}
    \caption{\footnotesize{Comparison of the color gradients seen when illuminating a curved sensor shape that is not radially symmetric around the vertical axis. The lack of blending colors can be seen on the left when only red, green, and blue light illuminates the sensor. A more gradual gradation of colors can be seen in the rainbow illuminated sensor.}}
    \vspace{-15pt}
    \label{fig:ellipsoid}
\end{figure}

To further evaluate the advantage of using the rainbow LED illumination, we run our calibration procedure on other lighting strategies previously introduced and commonly used in other curved tactile sensors. We compare the tactile images and resulting RMSE testing losses for the normal estimation network in Table 1. This includes the common red-green-blue illumination pattern introduced in \cite{johnson2009retrographic} and adopted in \cite{do2023densetact, azulay2023allsight, gomes2020geltip}, a single-colored light source \cite{lin2023dtact}, and the cross illumination scheme of GelSight360 \cite{tippur2023gelsight360}. We probe the sensor bodies using the lighting schemes described, train the gradient estimation network (with early stopping) for each, and compare their RMSE testing losses. The sensor shape with the cylindrical base and hemisphere top is chosen for testing since the sensor geometry allows for the 2-directional parallel lighting  discussed in \ref{sensor_illumination_strategy} on the sides of the sensor, and the traditional color gradient effect on the hemispherical portion of the sensor. The hardware and illumination methods of RainbowSight help it outperform GelSight360, which is an improvement considering there are no longer any occluded areas in the sensor. Additionally, the rainbow illumination scheme offers slightly better performance over the traditional red-green-blue pattern in estimating the gradients and uses fewer epochs to train. However, even though there is only a slight improvement in the accuracy of the gradient prediction for this shape, the rainbow illumination system can still be more advantageous in alternate curved sensor geometries, especially those that are not radially symmetric around the vertical axis. For example, in Figure \ref{fig:ellipsoid}, when the red-green-blue illumination pattern is used in the elliptical sensor shape, the harsh color directionality characteristic of the semi-specular coating is observed. In contrast, when illuminated with the rainbow illumination system, a more gradual color gradient is observed, causing more variation in RGB image intensities across the surface of the sensor. Features such as this may be useful in improving training time and accuracy for unexplored shapes.

\section{CONCLUSION}
In this work, we introduce a novel rainbow illumination method that can be implemented in a variety of shapes and sizes for curved, camera-based tactile sensors. Rather than illuminating the inside of the sensor with only red, green, and blue wavelengths, we employ an illuminated color gradient around the perimeter of the sensor body and photometric stereo techniques to produce 3D depth reconstructions of the deformation in the sensor surface when in contact with an object. This rainbow gradient is achieved by digitally varying the intensities of the red, green, and blue diodes in a ring of addressable RGB LEDs and is shown to produce satisfactory illumination for use with photometric stereo techniques with minimal hardware iterations or tuning. When compared to the previous cross-LED illumination strategy presented in \cite{tippur2023gelsight360}, we show improvements in ease of manufacturing, depth reconstruction results, and overall design generalizability to a wider range of sensor configurations, now including both half-sensor and omnidirectional sensor shapes. Additionally, we are able to eliminate the occlusions to the sensing area (previously caused by the LED cross in the sensor body) without sacrificing the quality of the depth reconstructions by housing the rainbow illumination source only at the sensor base. By designing this family of curved, rainbow-illuminated sensors that can be easily manufactured in a variety of shapes, provide good tactile data with minimal optical hardware tuning, and be fitted on a variety of robotic end effectors, we aim to make tactile sensing more accessible to roboticists as they pursue more complex manipulation tasks.





\section*{ACKNOWLEDGMENTs}

This material is based upon work supported by the National Science Foundation Graduate Research Fellowship, Toyota Research Institute (TRI), and the Amazon Science Hub. The authors would also like to thank Sandra Q. Liu and Arpit Agarwal for helpful discussions on sensor improvements and depth reconstruction methods.  

\bibliographystyle{IEEEtran}
\bibliography{citations}

\end{document}